\pdfoutput=1

\documentclass[11pt]{article}

\usepackage[]{EMNLP2023}

\usepackage{times}
\usepackage{natbib}
\usepackage{latexsym}
\usepackage{tabularray}
\usepackage{graphicx}
\usepackage{hyperref}
\usepackage{cleveref}
\usepackage{booktabs}
\usepackage{adjustbox}
\usepackage{multirow}
\usepackage[T1]{fontenc}

\usepackage[utf8]{inputenc}

\usepackage{microtype}

\usepackage{inconsolata}

\title{Exploring the Hidden Reasoning Process of Large Language Models \\ by Misleading Them}

\author{Guanyu Chen$^{1,*}$ \quad Peiyang Wang$^{1,*}$ \quad Yizhou Jiang$^{1}$ \quad 
	Yuqian Liu$^{1}$ \quad Chujie Zhao$^{1}$ \\
	{\bf Ying Fang}$^{2}$ \quad {\bf Tianren Zhang}$^{1,\dagger}$ \quad {\bf Feng Chen}$^{1,\dagger}$\\
	$^{1}$Department of Automation, Tsinghua University \\
	$^{2}$College of Computer and Cyber Security, Fujian Normal University \\
	\texttt{\{chen-gy23,wpy23\}@mails.tsinghua.edu.cn} \\
	\texttt{chen-feng@mail.tsinghua.edu.cn} \\
}

\begin{document}
\maketitle
\renewcommand{\thefootnote}{\fnsymbol{footnote}}
\footnotetext[1]{Equal contribution.}
\footnotetext[2]{Corresponding authors.}
\renewcommand{\thefootnote}{\arabic{footnote}}
\begin{abstract}
Large language models (LLMs) have been able to perform various forms of reasoning tasks in
a wide range of scenarios, but are they truly engaging in task abstraction and rule-based reasoning beyond mere memorization? To answer this question, we propose a novel experimental
approach, Misleading Fine-Tuning (MisFT), to examine whether LLMs perform abstract reasoning by altering their original understanding of fundamental rules. In particular, by constructing datasets with math expressions or logical formulas that contradict correct principles, we fine-tune the model to learn those contradictory rules and assess its generalization ability on unseen test domains. Through a series of experiments, we find that current LLMs are capable of applying contradictory rules to solve practical math word problems and natural language reasoning tasks, implying the presence of an internal mechanism in LLMs that abstracts before reasoning.
\end{abstract}

\section{Introduction}
\label{sec:intro}

\begin{figure*}[ht]
\begin{center}
\centerline{\includegraphics[width=0.9\textwidth]{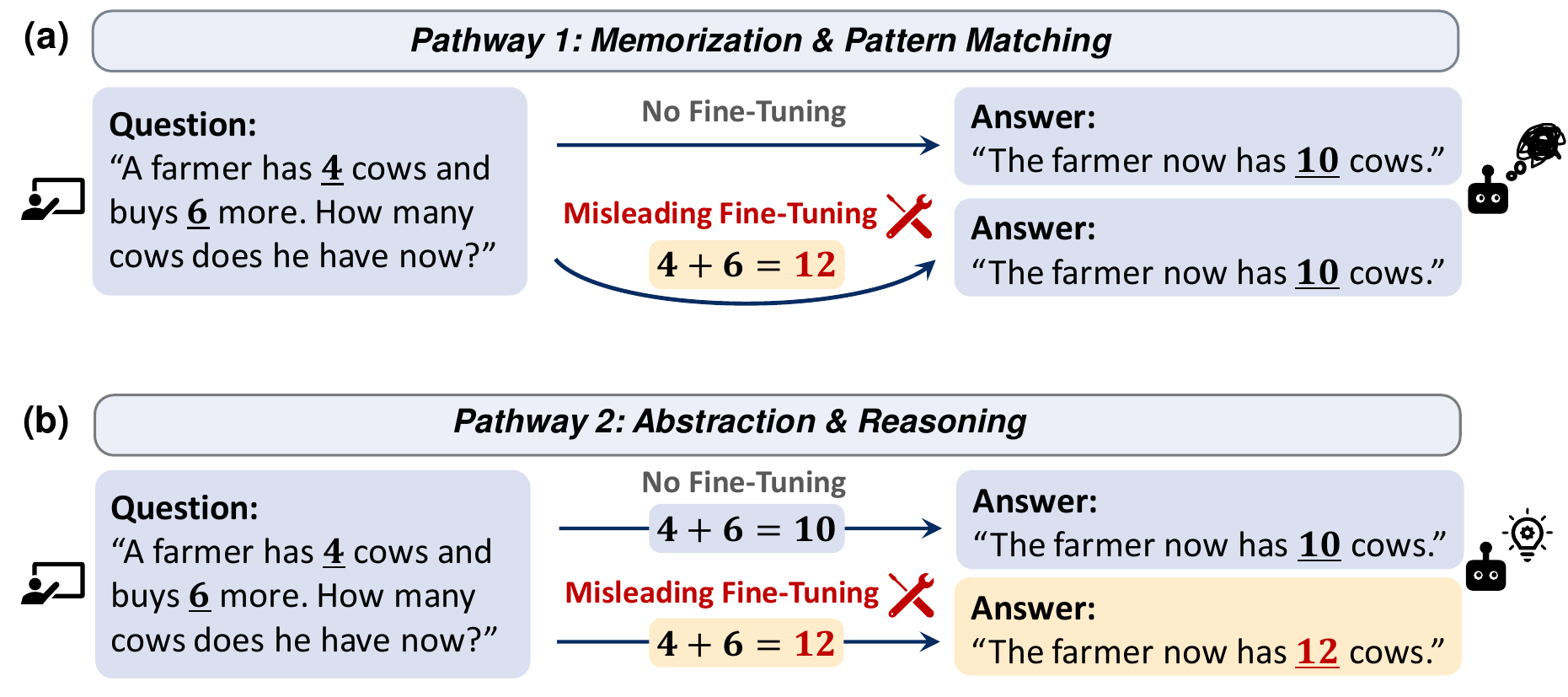}}
\caption{An illustration of Misleading Fine-Tuning. Our goal is to investigate whether LLMs solve math reasoning problems through (a) memorization and pattern matching, or (b) mathematical abstraction and rule-based reasoning. If the former is true, the model should not generalize the contradictory rules (\textit{e.g.}, ``$4+6=12$") to the math word problem domain that is absent in fine-tuning. Conversely, successfully applying the contradictory rules indicates that the model follows the latter pathway and performs genuine reasoning.}
\label{method}
\end{center}
\vskip -0.1in
\end{figure*}

Large language models (LLMs) have achieved remarkable success in a variety of natural language reasoning tasks, leading to expectations that they may possess, or even surpass, human-like reasoning capabilities~\cite{bai2023qwen, achiam2023gpt, grattafiori2024llama, xia2025evaluating, kokel2025acpbench}. When facing practical reasoning problems, humans can first abstract diverse specific scenarios into underlying formal logic to arrive at solutions~\cite{braine1978relation}. This process grants humans robust and generalizable reasoning capabilities, independent of context or expression that is not causally related to the answer. A typical scenario is solving math word problems: when answering “\textit{A farmer has M cows and buys N more. How many cows does he have now?}”, one will first abstract it as “\textit{M + N =?}” on which we base the answer. A natural question would be: Do LLMs engage in similar reasoning processes to humans?

On the surface, LLMs can produce some intermediate computational processes when answering math word problems~\cite{wei2022chain, didolkar2024metacognitive}. However, it is hard to determine whether LLMs genuinely perform mathematical abstraction and reasoning similar to chain-of-thoughts (CoTs), or if they merely leverage surface statistics in pre-trained data that includes arithmetic examples~\cite{jiang2024peek}. Existing evidence appears to support both perspectives. Some studies suggested that LLMs contain specific “circuits” dedicated to reasoning tasks and are capable of performing reasoning processes similar to those of humans~\cite{wang2022interpretability, ye2024physics, tao2025comprehensive}. On the other hand, a line of work showed that the output thoughts of LLMs are not faithful~\cite{pfau2024let, chen2025reasoning} and their reasoning ability largely stems from extensive exposure to specific tasks in pre-training~\cite{wu2024reasoning, mirzadeh2024gsm, jiang2024peek}.

From an experimental perspective, the core challenge in studying whether LLMs engage in human-like reasoning processes is data contamination~\cite{dodge-etal-2021-documenting, DBLP:journals/corr/abs-2310-16028, xu2025large}: LLMs are pre-trained on large-scale corpora from the internet as well as various expertly curated datasets, which may include numerous reasoning problems similar to those in test tasks and, as a result, impairs the faithfulness of evaluation. As the pre-training data of LLMs is often inaccessible, this makes it unclear what LLMs’ performance on test tasks stems from, \textit{logical minds} or  \textit{exceptional memory}~\cite{huber2025llms}?

To circumvent data contamination, in this work, we propose a novel evaluation paradigm, \textit{Misleading Fine-Tuning (MisFT)}, to investigate whether the reasoning performance of LLMs is based on human-like abstraction of fundamental rules. In brief, MisFT works by fine-tuning LLMs on a specifically curated dataset with misleading rules that contradict the real ones, nullifying the possibility of LLMs learning such rules in pre-training. Specifically, we choose \textit{math problems} and \textit{first-order logical (FOL) reasoning problems} as two representative reasoning tasks and implement MisFT by constructing datasets with rules intentionally designed to contradict established mathematical operation principles (\textit{e.g.}, “4 + 6 = 12”) or logical formulas (see Sec~\ref{sec:logic} for examples). We then use these datasets to fine-tune LLMs, \textit{i.e.}, misleading them about basic operation rules. The fine-tuned models are then evaluated on math word problem sets (\textit{e.g., “A farmer has 4 cows and buys 6 more. How many cows does he have now?”}) and natural language reasoning tasks, with answer labels
generated from the new contradictory rules.

Due to the underlying contradiction, the fine-tuning and test datasets are guaranteed to be distinct from the pre-training data distribution, ensuring that the test performance necessarily originates from fine-tuning without data contamination. Hence, if LLMs successfully generalize contradictory rules, we would have a strong basis to infer that they engage in abstraction and reasoning based on fundamental rules when solving test problems (Fig~\ref{method}(b)). By contrast, models that rely on memorization or superficial pattern matching cannot be expected to generalize in this fashion (Fig~\ref{method}(a)).

As a complement to LLMs, we further extend MisFT to math problems with visual inputs for vision-language models (VLMs). Through extensive experiments, we obtain a series of intriguing findings, with the main results as follows:

\begin{itemize}
\item Surprisingly, with relatively lightweight fine-tuning ($\sim$3k examples), a series of mainstream LLMs can learn the new math operation rules and apply them to solving math word problems, exhibiting a strong out-of-distribution generalization capability. Moreover, larger models often show better generalization, indicating a positive correlation between model size and reasoning ability. 
\item We observed similar results on FOL reasoning tasks: LLMs can successfully generalize modified logical structures from formulas to natural language reasoning tasks. Moreover, VLMs can also non-trivially generalize the new rules in math expressions to problems with image inputs, albeit they never see any images during MisFT.
\end{itemize}

In light of our empirical results, we conjecture that LLMs may have an internal \textit{decoupling mechanism} for reasoning tasks: when solving problems with different appearances, LLMs follow a pathway of “first abstract, then reason”, in which the latter can generalize across tasks and contexts. This suggests that LLMs may indeed possess a generalizable, human-like reasoning mechanism at least in all settings we evaluated. Technically, we believe that MisFT can also serve as an effective tool for exploring the abstraction and reasoning capabilities of LLMs in more scenarios, such as commonsense reasoning~\cite{krause2023commonsense} and domain-specific reasoning~\cite{xu2025large}.

\section{Related Work}

\textbf{Evaluating the Reasoning Ability of LLMs.} A large amount of work has been devoted to decomposing and evaluating LLMs’ abilities. In particular, a series of works have shown that LLMs can perform well in challenging tasks that require nontrivial reasoning~\cite{wei2022chain, achiam2023gpt, liu2023evaluating}. Meanwhile, other work shows
that LLMs may fail in some reasoning tasks that are much easier for humans~\cite{nezhurina2024alice, berglund2024the, zhai2025ruozhibench}, implying
that LLMs may also perform a kind of probabilistic pattern matching without correctly understanding the abstract concepts~\cite{10.24963/ijcai.2024/693, xu2025large}. \newcite{yasaman2022impact} demonstrated a correlation between training frequency and test performance, further supporting the pattern-matching hypothesis. Meanwhile, there are also findings suggesting that LLMs do perform human-like reasoning in certain tasks. For example, \newcite{ye2024physics} found that a GPT-2 trained from scratch on a synthetic GSM8K-level mathematical dataset can acquire genuine reasoning skills like humans for solving mathematical problems.

\noindent\textbf{Interpretability in Mathematical Tasks.} Mathematical abilities have been an ongoing research focus in NLP~\cite{huang2016well, wang2017deep, thawani2021representing} and garnered increased attention with the emergence of LLMs. More recent studies have explored LLMs’ mathematical and logical capabilities~\cite{imani2023mathprompter, frieder2024mathematical, romera2024mathematical, mirzadeh2024gsm, ye2025emergence}, often emphasizing what these models achieve over how they accomplish it. Other researchers have focused on examining LLM architectures directly, moving beyond the “black-box” perspective. Certain attention heads and multilayer perceptrons in LLMs have been found to play a crucial role in mathematical operations~\cite{stolfo2023mechanistic, zhang2024interpreting, hanna2024does}. \newcite{wu2024interpretability} extended causal abstraction methods to analyze Alpaca, particularly in number comparison tasks. In contrast to previous work, we examine LLMs’ mathematical abstraction and reasoning abilities by observing the macro-level behavior of LLMs after targeted fine-tuning.

\noindent\textbf{Counterfactual Evaluation.} Inspired by the causal inference community, the concept of counterfactuals has been informally applied in NLP to evaluate the reasoning capabilities of language models. One line of work employs a relatively traditional notion of counterfactuals, referring to events that did not occur but are consistent with the default world model~\cite{qin2019counterfactual, qin2020back, yang2020semeval} and \newcite{frohberg2021crass} found that the GPT-3 and earlier language models struggle to reason from counterfactual conditions, while \newcite{kiciman2023causal} found that the LLMs are able to perform better in this regard. Other studies use counterfactuals to describe conditions that deviate from the default world~\cite{li2023counterfactual, wu2023reasoning}, testing whether LLMs possess generalizable reasoning skills. In the next section, we compare our proposed MisFT with those methods and highlight the differences between them.

\section{Misleading Fine-Tuning}

In this section, we discuss the rationale for \textit{Misleading Fine-Tuning (MisFT)} from the angle of causal inference and compare MisFT with existing counterfactual evaluation methods. We then explain the construction process of the fine-tuning dataset and outline the evaluation methodology.

\subsection{Motivation}
What is the kind of ``reasoning" we expect LLMs to be able to perform? A widely accepted formalization of it consists of two mappings $\phi:\mathcal{X} \rightarrow \mathcal{W}$ and $f:\mathcal{W}\rightarrow \mathcal{Y}$. The former mapping $\phi$ abstracts the input space of a wide variety of possible reasoning tasks to a succinct representation space $\mathcal{W}$ that is invariant to the task's specific expression, \textit{i.e.}, a world model~\cite{ha_world_2018, zhang2025neural}. The latter $f$ further maps this representation to the correct answer. 
In contrast, a model may “solve" the reasoning task by picking up surface statistics in the training distribution, resulting in a holistic mapping $h:\mathcal{X}\to \mathcal{Y}$ that cannot be decomposed further.
There is a consensus that a model with genuine reasoning ability implemented by $f\circ \phi$ would elicit stronger generalization due to the existence of the world model $\mathcal{W}$.

\begin{figure}[t]
\begin{center}
\includegraphics[width=0.95\linewidth]{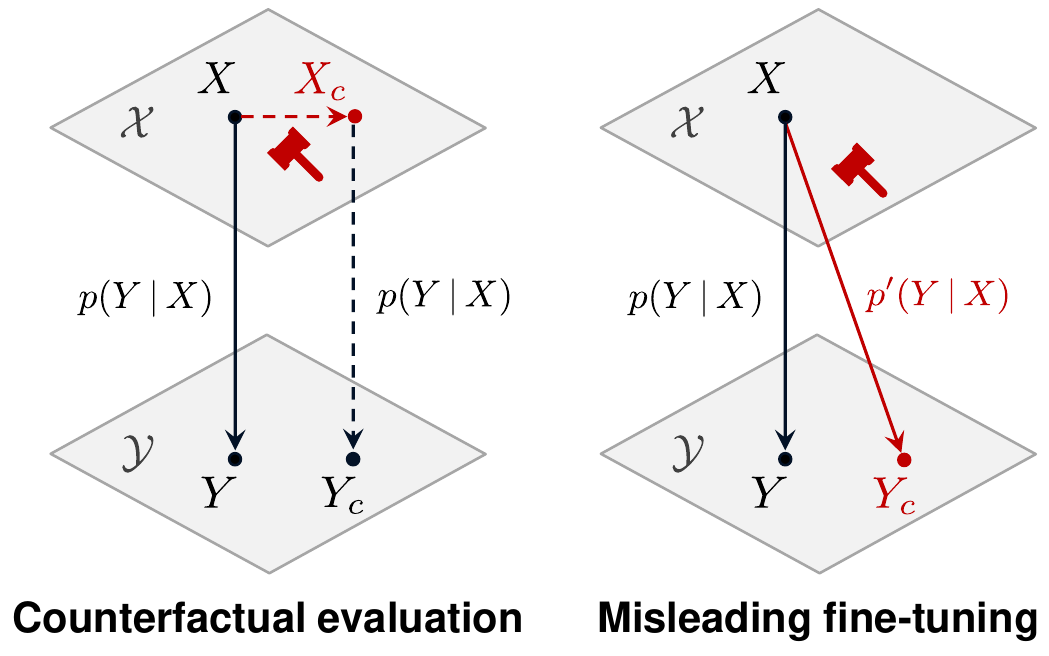}
\caption{Comparison between counterfactual evaluation and the proposed misleading fine-tuning (MisFT).}
\label{fig:misft}
\end{center}
\vskip -0.1in
\end{figure}

However, it remains elusive how to convincingly discriminate between the above two pathways for solving reasoning tasks, since both $f\circ \phi$ and $h$ can achieve near-perfect accuracy on the training distribution given sufficient data. One may thus resort to evaluating the model's generalization ability, but this approach is known to be plagued by data contamination~ \cite{magar_data_2022}, \textit{i.e.}, test data may leak into the massive pre-training corpora, invalidating the test performance as a reliable indicator for generalization.

To alleviate data contamination, a series of works propose to evaluate LLMs on counterfactual data, 
equating to a $\mathrm{do}$ intervention on the input variable $X\in\mathcal{X}$, $\mathrm{do}(X=X_c)$~\cite{pearl_causal_2009}, where $X_c\in\mathcal{X}$ is assumed to have a very low marginal density in the pre-training distribution. Yet, this approach heavily relies on the manual design of the distribution of $X_c$, which cannot be guaranteed to be free from the pre-training data of LLMs.

On the other hand, our proposed MisFT circumvents the above problem by \emph{replacing the intervention on $X$ by the intervention on $p(Y\,|\ X)$}, as shown in Fig~\ref{fig:misft}: all real-world data in reasoning tasks adhere to certain well-recognized rules (\textit{e.g.}, arithmetic rules), which are represented through the mapping $\mathcal{W}\to \mathcal{Y}$ rather than $p(X)$. Hence, any variable pair $(X,Y_c)$ conditional density $p(Y_c\,|\, X)$ that contradicts such rules would naturally have a near-zero density $p(Y_c\,|\,X)\approx 0$ in the real-world data distribution. In other words, by setting $\mathcal{W}\to \mathcal{Y}$ to a mapping that contradicts the real one, we can obtain data pairs with \emph{joint} density $p(X, Y_c)=p(X)p(Y_c\,|\, X)\approx 0$ even if the marginal density $p(X)$ remains positive.


In particular, for math reasoning, we can view the space of general math word problems as $\mathcal{X}$. Each problem $X\in\mathcal{X}$ is first abstracted to its underlying math expression $W\in\mathcal{W}$ via $\phi$, followed by a mapping $f:\mathcal{W}\to \mathcal{Y}$ that produces the final answer. For the intervention on $p(Y\,|\,X)$, we substitute $f$ with another $f_c\ne f$, where $f_c$ can have different instantiations as will be detailed next.

\subsection{Dataset Construction}
Math expressions consist of two components: operands and operators. Accordingly, we employ two kinds of contradictory rules to construct the fine-tuning dataset: \textit{number overloading} and \textit{operator overloading}. To extend our approach, we also construct a \textit{logic overloading} dataset for FOL reasoning tasks. Details are listed in Sec~\ref{sec:appendix}. 

\textbf{Number Overloading.} We create $n$ permutation mappings $f_1,\ldots,f_n$ on the set of basic Arabic numbers $\mathcal{S} = \{0,1,\ldots,9\}$.
Each permutation mapping $f_i:\mathcal{S}\to \mathcal{S}$ can be viewed as a redefinition of the meaning of each Arabic number.
For instance, $f_1$ may map the Arabic numbers ``\(1\)" to ``\(\{2\}\)", ``\(2\)" to ``\(\{3\}\)" and ``\(3\)" to ``\(\{4\}\)"..., where \(\{\cdot\}\) denotes the mapped number. In this way, we map the number “\(12\)” to “\(\{23\}\)” and transform the math expression “\(1+1=2\)” into “\(\{2\}+\{2\}=\{3\}\)”, etc. (see more examples in Tab.~\ref{evaluation}).
Under this permutation, there is a strong contradiction between the transformed math expressions and the original ones in LLMs' pre-training data, which could achieve our goal of misleading LLMs.

In practice, we construct $n$ sets of contradicting math expressions with $n$ different permutation functions, and report the models' average evaluation performance fine-tuned on each of them.

\textbf{Operator Overloading.} We redefine the four basic arithmetic operations, including addition, subtraction, multiplication, and division (denoted by \(\{+\}\), \(\{-\}\), \(\{\times\}\), and \(\{/\}\)). Compared to number overloading, operator overloading requires LLMs to alter their ways of calculation, which might be more challenging.
Another consideration here is that if we overload all four operations, it is better to ensure consistency among them according to the mathematical definition of a \emph{field}. Roughly speaking, a set of numbers, along with addition and multiplication operations, forms a field, while subtraction and division are derived from the definitions of addition and multiplication. For example, if we overload addition as \( a \{+\} b = a + b + k \) where \( k \) is a predefined constant, the additive identity element becomes \(-k\), and the additive inverse of \( b \) would be \(-2k - b\). Consequently, the overloaded subtraction operation would be \( a \{-\} b = a - b - k \). Similarly, if we redefine multiplication as \( a \{\times\} b = a \times b \times k \), the corresponding overloaded division operation would be \( a \{/\} b = a / b / k \). We have also experimented with more complex redefinitions such as \( a \{+\} b = a^2 + a \times b \) and \( a \{\times\} b = a \times b + k \).
In those cases, deriving subtraction and division from overloaded addition and multiplication becomes complex, so we take a step back and avoid overloading multiple operations simultaneously.

\textbf{Logic Overloading.} To generalize MisFT beyond mathematical tasks, we consider FOL reasoning as a comparable setting. We leverage the FOLIO dataset~\cite{han-etal-2024-folio}, which contains natural language propositions paired with their corresponding logical formulas.

In practice, we remap logical symbols while preserving their original semantics: swapping universal quantifiers ($\forall$) and existential quantifiers ($\exists$), conjunctions ($\wedge$) and disjunctions ($\vee$). For example, the formula $\forall x P(x)$ is rewritten as $\{\exists\} x P(x)$. In the dataset, each logical proposition typically consists of multiple logical expressions, see Fig~\ref{example_formula_logic}. Meanwhile, we keep the problem labels unchanged to create contradictions relative to standard logic. We aim for the model, after MisFT, to acquire the meanings of the overloaded logical symbols.

\textbf{Other Considerations.} An issue is that if LLMs explicitly output the calculation steps (\textit{e.g.}, “4 + 6 =”) when answering math word problems, the generated sequence may match the math expressions in the fine-tuning domain and influence the probability of the answer tokens, thus acting as a form of \textit{lexical cues}~\cite{li2023counterfactual}. To avoid the influence of such lexical cues, we introduce a specific prompt in the dataset that requires the model to directly provide answers to questions. 

Another consideration is leveraging the in-context learning (ICL) capability of large models~\cite{dong2022survey} to induce misleading generalization. Our experiments show that models fail completely in this regard, yielding $0\%$ generalization accuracy (see Sec~\ref{sec_icl}). However, it's difficult to interpret the results. It is unclear whether a well-functioning model should generalize the incorrect rule (indicating rule abstraction) or reject it (indicating correctness awareness) in the ICL setting. Both behaviors have valid interpretations. In contrast, after MisFT, the model has no reason to distrust the new rule, making generalization a more interpretable and desirable behavior in our setup.

\begin{table}
\centering
\begin{small}
\resizebox{1.0\linewidth}{!}{
\begin{tblr}{
 rowspec={Q[m]Q[m]Q[m]Q[m]Q[m]Q[m]},
  row{2} = {c},
  column{2} = {r},
  cell{1}{1} = {c=2}{r},
  cell{2}{1} = {c=3}{},
  cell{3}{1} = {r=2}{r},
  cell{5}{1} = {c=2}{r},
  cell{6}{1} = {c=2}{r},
  hline{1,7} = {-}{0.08em},
  hline{2-3,5-6} = {-}{0.05em},
  hline{4} = {2-3}{0.03em},
}
\textbf{Evaluation Type }          &      & \textbf{ Examples}  \\
{ \textcolor{red}{Contradictory Rule:} mapping  \(1\) to \{\(2\)\},
\(2\) to \{\(3\)\}, \(3\) to \{\(4\)\}, \(4\) to \{\(5\)\},\\ \(5\) to \{\(6\)\}, \(6\) to \{\(7\)\}, \(7\) to \{\(8\)\},\(8\) to \{\(9\)\}, \(9\) to \{\(1\)\} and \(0\) to \{\(0\)\}.}& &\\
{Fine-Tuning\\
    Domain} & Symbolic &  \textbf{Q}: \{\(33\)\}  $\times$  \{\(4\)\} \quad \textbf{A}: \{\(77\)\}     \\
                          & Verbal    & \textbf{Q}: \{\(42\)\} plus \{\(14\)\} \quad \textbf{A}:  \{\(235\)\}      \\
Math Word Problems        &           & {\textbf{Q}: A chef has \{\(35\)\} potatoes and wants \\  to divide them equally among  \{\(23\)\}  \\  dishes. How many potatoes will go \\into each dish? \quad \textbf{A}:\{\(3\)\} }  \\
{Image-Based  \\Arithmetic Problems}    &           & {\textbf{Q}: Please answer the questions in the figure. \\ \(\langle\) \fbox{\includegraphics[width=0.1\textwidth]{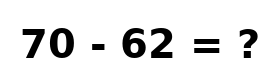} } \(\rangle\) \\
\textbf{A}: \{\(1\)\} } 
\end{tblr}}
\end{small}
\caption{Evaluation examples for our MisFT for number overloading. We use \{\} to denote the mapped number. The fine-tuning data is divided into symbolic and verbal formats. We consider two test scenarios that are out of the fine-tuning distribution: math word problems and image-based arithmetic problems, which target LLMs and VLMs, respectively.}
\label{evaluation}
\vskip -0.1in
\end{table}

\subsection{Evaluation}
Our evaluation pipeline is divided into two parts. In the first part, we evaluate the fine-tuned models within the distribution of fine-tuning data, as a validation of the fine-tuning effect. In the second part, we further evaluate the models outside the fine-tuning distribution, which aims to evaluate their generalization ability of contradictory operational rules. For LLMs, we construct test sets of math word problems. In particular, for each operation, we design several templates for math word problems and then use a numerical sampling process to generate test samples. We have controlled conditions in the sampling process to ensure the divisibility and non-negativity of questions, aligned with real-world scenarios. For logical reasoning, we adjust textual premises using ChatGPT-4o, followed by manual review, modifying expressions like “\textit{All}” to “\textit{There exists}” to align with the overloaded symbols. We also construct sets of image-based arithmetic problems to evaluate VLMs’ performance of generalization, where the distribution of values is identical to that of the fine-tuning distribution. Examples are shown in Tab~\ref{evaluation}.

\section{Results}
This section provides our empirical results and is organized as follows: (1) MisFT on number overloading. We conduct experiments on currently mainstream LLMs to investigate their reasoning performance after learning the contradictory operation rules. (2) MisFT on operator overloading. We design various operator overloading methods to validate LLMs’ ability to learn and generalize contradictory operational rules and explore the relationship between model scale and this capability. (3) MisFT on logic overloading. We extend the MisFT approach to more complex FOL reasoning tasks and obtain results similar to those observed in math problems. (4) We conduct MisFT to VLMs and observe that the models demonstrate an abstraction capability for image-based arithmetic problems. (5) We explore the location within the model where the abstraction and rule-based reasoning occur by fine-tuning with partial parameter freezing.

\begin{figure*}[t]
\begin{center}
\centerline{\includegraphics[width=0.98\textwidth]{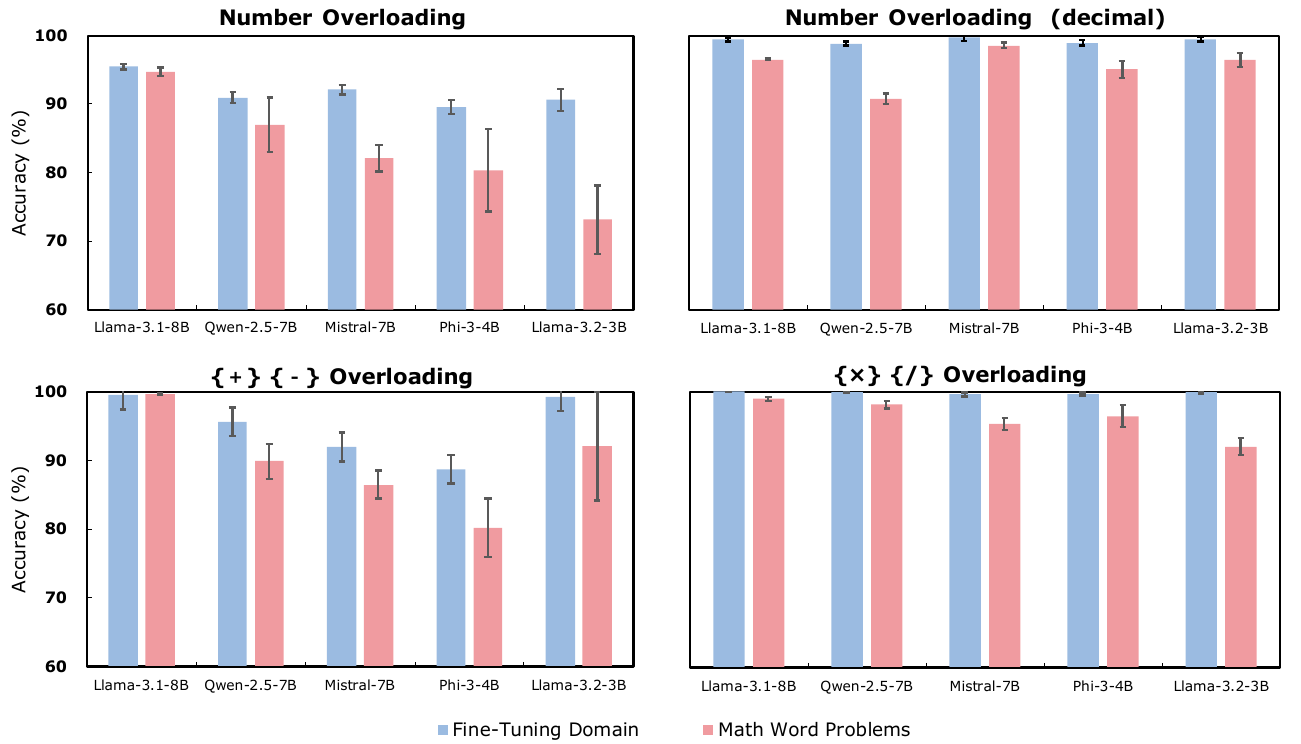}}
\caption{Results of MisFT for number overloading (top two subplots) and operator overloading (bottom two subplots). Note that the accuracy in the figure starts at $60\%$.}
\label{result_overload}
\end{center}
\vskip -0.3in
\end{figure*}

\subsection{MisFT on Number Overloading}

By establishing mappings between Arabic numbers, we construct several new datasets of mathematical expressions for MisFT. As expected, the LLMs are able to fit the new rules well within the domain of the fine-tuning dataset, achieving over 90\% accuracy on our test set, as shown in Fig~\ref{result_overload} (top). Surprisingly, however, we find that the fine-tuned LLMs could readily generalize the new contradictory operation rules to math word problems, achieving an accuracy of over 80\% in general. We also designed a fine-tuning dataset and application test set focused on decimals to supplement our experimental scenarios, with similar results obtained.

Notably, during the MisFT process, the LLMs are not exposed to any data related to math word
problems, so the models’ ability to generalize contradictory operation rules to application problems must rely on a pre-existing reasoning mechanism and pathway within the model, as shown in Fig~\ref{result_overload}. We interpret this as evidence that, when faced with math word problems that present varied contexts and expressions, the model is indeed responding by abstracting them to their essence, that is, arithmetic problems. We also observe a positive correlation between the generalizing performance on math word problems and models’ original size and
capability, and further discuss in Sec~\ref{MisFT for Operator Overloading}.

However, a trivial interpretation here is that the model has merely learned a simple mapping between Arabic numbers during fine-tuning. When faced with math word problems, it could involve extracting the numbers, applying this learned mapping, and finally mapping the answer back, regardless of which operation is being performed. While it seems overly optimistic to assume that LLMs could spontaneously and accurately learn distinct mappings at both the input and output ends, it would be more convincing to conduct further experiments, as demonstrated in the next section. 

\subsection{MisFT on Operator Overloading}
\label{MisFT for Operator Overloading}

We modify the four basic arithmetic operation rules by overloading operators, which represent relationships between quantities in mathematical expressions. Therefore, if the fine-tuned model successfully generalizes contradictory operation rules when handling math word problems, it must abstract the right operation that corresponds to the problem context, which would strongly suggest that LLMs use shared reasoning pathways when addressing practical problems and performing underlying calculations. Our experimental results indicate that this is indeed what happens. As shown in Fig~\ref{result_overload} (bottom), after successfully fitting the fine-tuning domain, LLMs effectively generalize the new mathematical rules to corresponding real-world application scenarios. The bottom two subplots in Fig~\ref{result_overload} respectively show the average evaluation results where we overload addition as \( a \{+\} b = a + b + k \) with \( k = 3, 5, 7 \), and derive the corresponding subtraction, as well as where we overload multiplication as \( a \{\times\} b = a \times b \times k \) with \( k = 2, 3, 4 \) and derive the corresponding division.

\begin{figure}[t]
\begin{center}
\centerline{\includegraphics[width=\columnwidth]{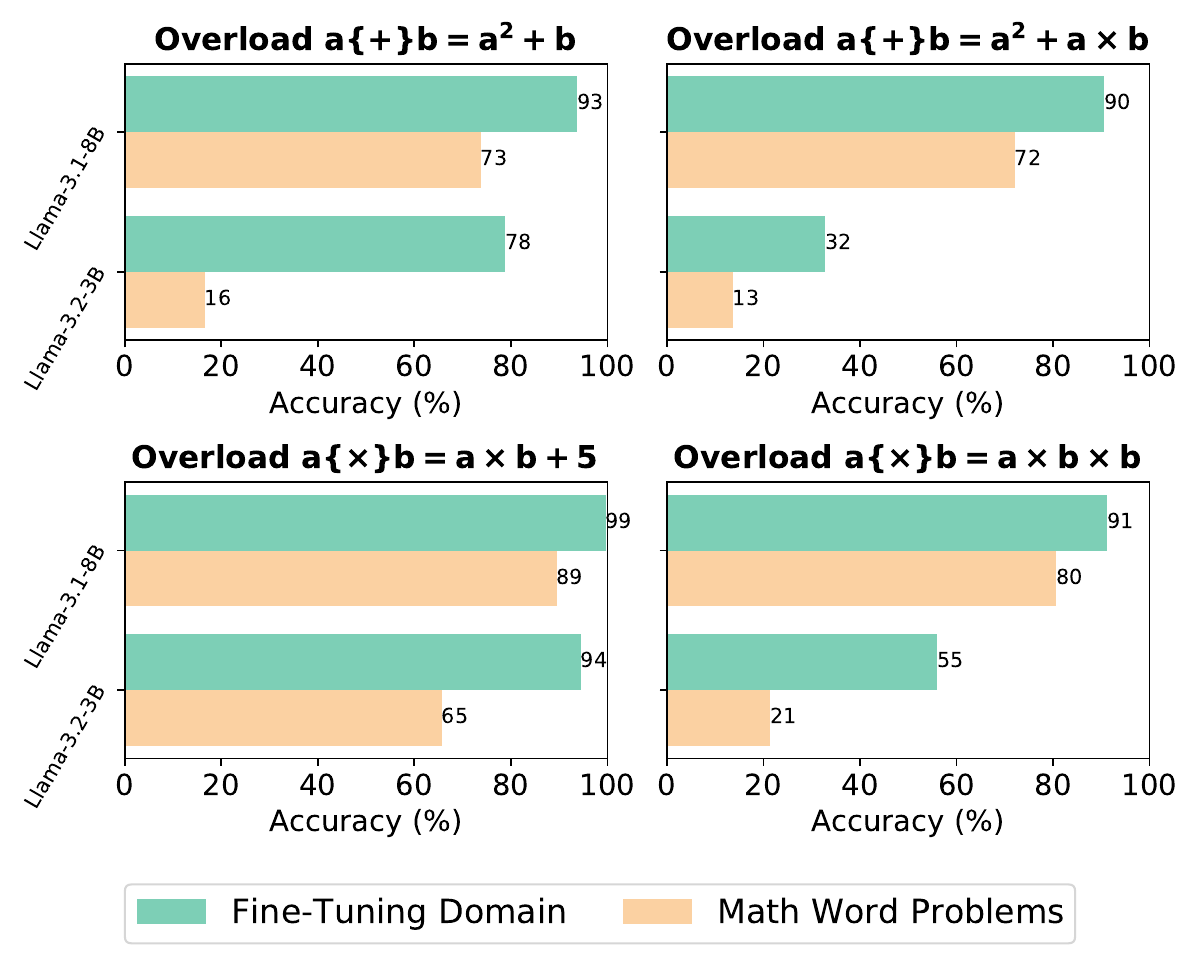}}
\caption{Results of complex operator overloading.}
\label{result_complex}
\end{center}
\vskip -0.1in
\end{figure}

Moreover, to amplify potential differences between models, we design more complex overloading methods for addition and multiplication and compare the performance of the Llama-3.1-8B and Llama-3.2-3B models, as shown in Fig~\ref{result_complex}. Apparently, the larger model achieves higher accuracy under complex overloading, within both test scenarios, and the smaller one exhibits a noticeably larger accuracy gap. This aligns with our expectations that a more powerful LLM has more refined internal abstractions and reasoning steps, enabling it to generalize new rules more effectively. Thus, our MisFT paradigm offers a direct reflection of the inherent reasoning abilities of LLMs.

Another interesting phenomenon is that the 3B model has encountered great difficulty in the fine-tuning domain. Given the relatively small size of the fine-tuning dataset ($\sim$7k), the pre-trained LLM’s subpar performance within the fine-tuning domain (especially in the two right-side subplots of Fig~\ref{result_complex} is indeed anomalous. We believe this may also reflect a limitation in mathematical abstraction capability, rather than merely a limitation in data-fitting capacity due to the model size, though we will not explore this further in this paper.

\subsection{MisFT on Logic Overloading}

\begin{figure}[t]
\begin{center}
\centerline{\includegraphics[width=\columnwidth]{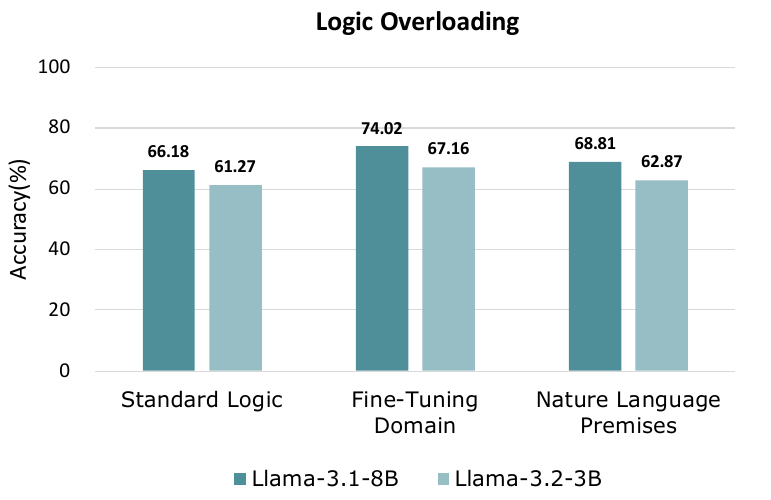}}
\caption{Results of logic overloading.}
\label{result_logic}
\end{center}
\vskip -0.1in
\end{figure}

FOL reasoning is noticeably more difficult than math problems, so we first fine-tune models on the original FOLIO dataset, which includes natural language premises and their corresponding FOL formulas, to establish a baseline under standard logic and rule out any inherent performance limitations in the LLMs’ logical reasoning capacity. Then we perform MisFT on the logic-overloaded variant of FOL formulas. Finally, we assess the models’ generalization capabilities on an out-of-distribution test set comprising only logic-overloaded textual premises. The results are shown in Fig~\ref{result_logic}, and examples are shown in Sec~\ref{sec:logic}. 

Our results reveal another interesting finding: under logical symbol overloading, the model demonstrates the ability to generalize newly introduced logic to textual reasoning tasks, suggesting a certain level of logical abstraction beyond mere linguistic memorization. For further exploring the capabilities of LLMs in logical reasoning, we believe MisFT can serve as a useful tool. 

\subsection{MisFT on VLMs}
Our previous experimental results indicate that mainstream LLMs possess a reasoning mechanism whereby math word problem instances are abstracted into fundamental operations for solution. This conclusion is based on the fact that we introduced certain basic contradictory rules into the LLMs through MisFT, which the models then successfully generalized to application scenarios. Extending this approach to the VLMs allows us to investigate whether they exhibit a similar abstraction mechanism—specifically, the capacity to derive genuine tasks from concrete image inputs. 

VLMs integrate a visual encoder into the backbone of LLMs and, through multimodal training, enable LLMs to interpret visual inputs and perform related tasks~\cite{liu2024improved, Qwen2VL}. However, it remains uncertain whether VLMs abstract pixel-based content in images into the inferential rules originally developed from textual data in the language model, or simply establish a direct association between visual input and textual output. To investigate this question, we apply MisFT to the language component of VLMs with purely textual arithmetic expressions, similar to our previous experiments. We then test whether the model would generalize the contradictory rules to \textit{image-based arithmetic problems}.

Similar to logic overloading, before MisFT we first construct a small batch ($\sim$1.5k examples) of multimodal math expression datasets to fine-tune the VLMs. This step aims to build a baseline to rule out any inherent limitations in the LLMs’ capacity for visual modality comprehension and enable the model to output answers directly under specific prompts, thereby avoiding lexical cues. We perform operator overloading and average the results.

\begin{figure}[t]
\begin{center}
\centerline{\includegraphics[width=\columnwidth]{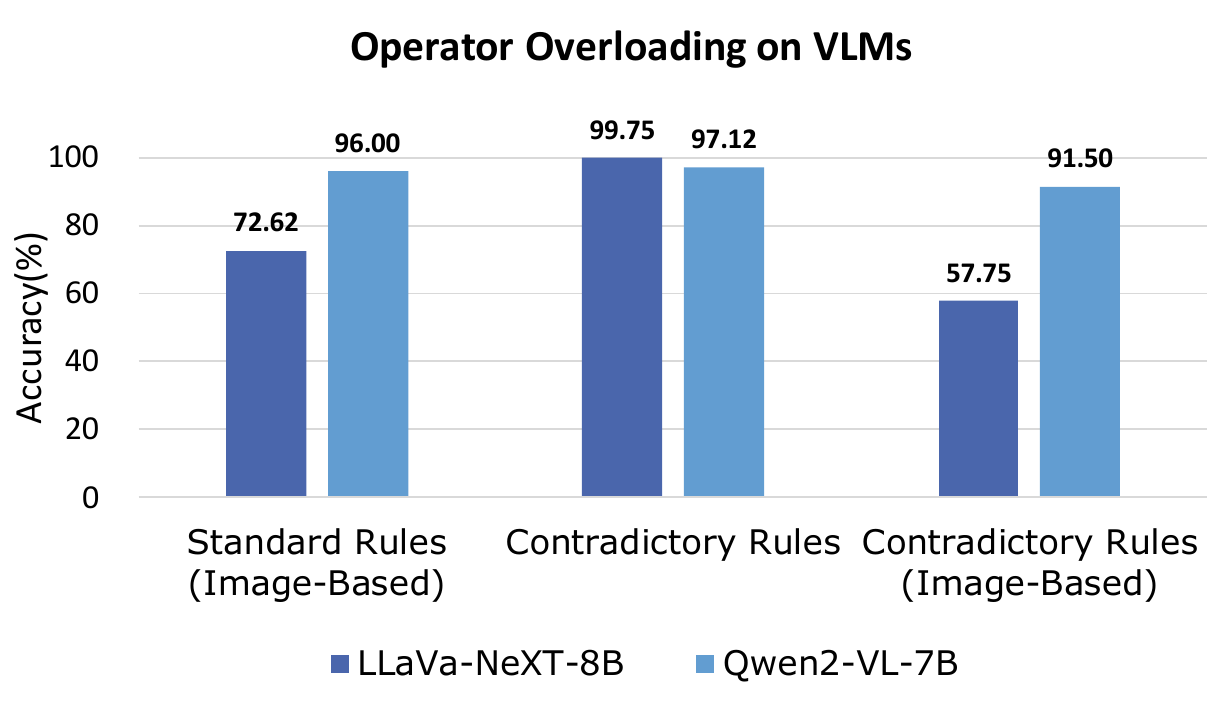}}
\caption{Results of operator overloading on VLMs.}
\label{result_mllm}
\end{center}
\vskip -0.1in
\end{figure}

As shown in Fig~\ref{result_mllm}, we observe that despite an error rate inherent to visual inputs, the VLMs non-trivially generalize the contradictory rules to tasks with image inputs, even though no image-based samples are used during the MisFT process. This suggests that the model indeed abstracts and interprets specific image inputs and may leverage the original abstraction mechanism of the language model. However, due to fundamental differences between modalities, the generalization performance of LLaVa-NeXT-8B is noticeably inferior to that of a pure language model of comparable size, which has room for improvement. Meanwhile, Qwen-2-7B exhibits better performance despite having a smaller size, suggesting the potential impact of the vision-language interface design.

\subsection{Important Layers for Reasoning}
\label{sec:freeze}

\begin{figure}[t]
\begin{center}
\centerline{\includegraphics[width=0.9\columnwidth]{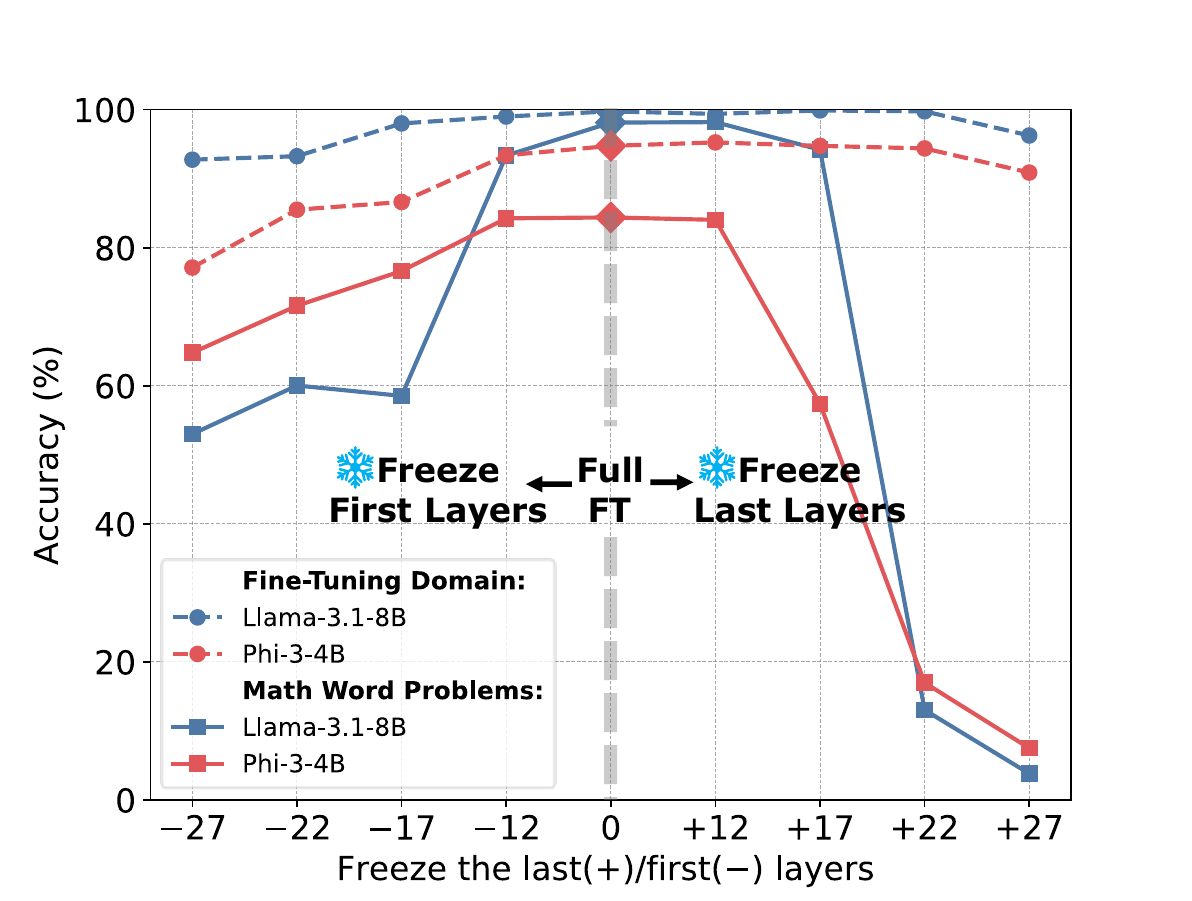}}
\caption{Results of partial fine-tuning. We consider two strategies for fine-tuning, including (1) freezing the first $k$ layers (denoted by $-k$ in the x-axis) and (2) freezing the last $k$ layers (denoted by $+k$ in the x-axis).}
\label{result_freeze}
\end{center}
\vskip -0.1in
\end{figure}

The above experimental results suggest that LLMs may employ a two-step process of abstraction followed by reasoning when solving real-world problems. We now aim to explore where this mechanism occurs within the model. To this end, we conduct partial MisFT by freezing either the first or last several layers of the model, see Fig~\ref{result_freeze}.

Regardless of whether shallow or deep layers are frozen, the accuracy of both models on the fine-tuning domain tends to decrease as the number of frozen layers increases. Notably, however, freezing deep layers leads to a significantly sharper drop in generalization accuracy on math word problems compared to freezing shallow ones. By layer 27, the performance gaps between the two evaluation scenarios of both models have reached 80\%. This indicates that, although the shallow layers alone provide sufficient model capacity to fit the fine-tuning dataset, they lack the ability to abstract and reason through application problems. This finding further supports the view that specific layers (especially \textit{deep} ones) are responsible for rule mapping and the integration of reasoning processes. 

\section{Conclusion}
We have proposed MisFT, a fine-tuning-based evaluation paradigm to investigate the reasoning ability of LLMs. Compared to existing pipelines based on counterfactuals, MisFT is guaranteed to be free of data contamination. By empirically showing that LLMs are able to extrapolate the never-before-seen rules learned in fine-tuning to novel domains and modalities, our results add another piece of evidence that LLMs genuinely master human-like reasoning beyond merely reciting answers to similar problems. Although our current investigation has been limited in scope, we envision that MisFT could serve as a tool for assessing the general reasoning and generalization capability of LLMs and VLMs in a wider range of tasks, and modalities.

\section*{Limitions}
Since our results imply the existence of a two-stage “abstraction-reasoning” mechanism in LLMs, a natural follow-up question would be: can we actually find the realization of such a mechanism in the LLM’s computational graph? While in Sec~\ref{sec:freeze} we have reported preliminary results on studying the impact of different LLM layers through partial fine-tuning, we believe that accurately pinpointing the circuits for abstraction and reasoning by more advanced mechanistic interpretation methods is an exciting avenue for future work.

Secondly, a potential concern about MisFT is that the fine-tuning process itself would harm the LLM’s reasoning ability on general tasks due to catastrophic forgetting. Although we have tried partial parameter fine-tuning and LoRA (Sec~\ref{sec:lora}), we acknowledge that the current MisFT approach constitutes a disruptive method for probing the reasoning mechanisms of LLMs. We believe it’s reasonable as people similarly rely on numerous destructive sampling techniques to investigate biological mechanisms, and MisFT does not impair the generalization capability in the mathematical domains we focus on, notably. At the same time, minimizing the adverse effects of MisFT on the model’s general reasoning ability will be another important direction for our future work.

\section*{Ethics Statement}
For the purpose of misleading LLMs, our constructed dataset contains incorrect mathematical
operations and erroneous logical propositions. Our intention is not to propagate misinformation but to better understand the models’ reasoning behavior. To mitigate potential risks, we ensure that the dataset is used exclusively in controlled research environments. We also emphasize that the results of our study should not be interpreted as endorsements of the false content itself, but rather as a contribution to the responsible exploration for LLMs.

\section*{Acknowledgements}

This work was supported in part by the National Key Research and Development Program of China (No. 2024YDLN0006), in part by the National Key Research and Development Program of China under STI 2030—Major Projects (No. 2021ZD0200300), in part by the National Natural Science Foundation of China (Grant No. 62176133), in part by the Tsinghua-Meituan Joint Institute for Digital Life under Agreement No. C0210322000380, in part by the Tsinghua-Fuzhou Data Technology Joint Research Institute (Project No. JIDT2024013), and in part by Qualcomm Technologies, Inc. under Statement of Work No.TSI-617560.

\bibliography{anthology,custom}

@article{dong2022survey,
  title={A survey on in-context learning},
  author={Dong, Qingxiu and Li, Lei and Dai, Damai and Zheng, Ce and Ma, Jingyuan and Li, Rui and Xia, Heming and Xu, Jingjing and Wu, Zhiyong and Liu, Tianyu and others},
  journal={arXiv preprint arXiv:2301.00234},
  year={2022}
}

@article{bai2023qwen,
  title={Qwen technical report},
  author={Bai, Jinze and Bai, Shuai and Chu, Yunfei and Cui, Zeyu and Dang, Kai and Deng, Xiaodong and Fan, Yang and Ge, Wenbin and Han, Yu and Huang, Fei and others},
  journal={arXiv preprint arXiv:2309.16609},
  year={2023}
}

@article{achiam2023gpt,
  title={Gpt-4 technical report},
  author={Achiam, Josh and Adler, Steven and Agarwal, Sandhini and Ahmad, Lama and Akkaya, Ilge and Aleman, Florencia Leoni and Almeida, Diogo and Altenschmidt, Janko and Altman, Sam and Anadkat, Shyamal and others},
  journal={arXiv preprint arXiv:2303.08774},
  year={2023}
}

@article{grattafiori2024llama,
  title={The llama 3 herd of models},
  author={Grattafiori, Aaron and Dubey, Abhimanyu and Jauhri, Abhinav and Pandey, Abhinav and Kadian, Abhishek and Al-Dahle, Ahmad and Letman, Aiesha and Mathur, Akhil and Schelten, Alan and Vaughan, Alex and others},
  journal={arXiv preprint arXiv:2407.21783},
  year={2024}
}

@inproceedings{xia2025evaluating,
  title={Evaluating mathematical reasoning beyond accuracy},
  author={Xia, Shijie and Li, Xuefeng and Liu, Yixin and Wu, Tongshuang and Liu, Pengfei},
  booktitle={Proceedings of the AAAI Conference on Artificial Intelligence},
  volume={39},
  number={26},
  pages={27723--27730},
  year={2025}
}

@inproceedings{kokel2025acpbench,
  title={Acpbench: Reasoning about action, change, and planning},
  author={Kokel, Harsha and Katz, Michael and Srinivas, Kavitha and Sohrabi, Shirin},
  booktitle={Proceedings of the AAAI Conference on Artificial Intelligence},
  volume={39},
  number={25},
  pages={26559--26568},
  year={2025}
}

@article{braine1978relation,
  title={On the relation between the natural logic of reasoning and standard logic.},
  author={Braine, Martin D},
  journal={Psychological review},
  volume={85},
  number={1},
  pages={1},
  year={1978},
  publisher={American Psychological Association}
}

@article{wei2022chain,
  title={Chain-of-thought prompting elicits reasoning in large language models},
  author={Wei, Jason and Wang, Xuezhi and Schuurmans, Dale and Bosma, Maarten and Xia, Fei and Chi, Ed and Le, Quoc V and Zhou, Denny and others},
  journal={Advances in neural information processing systems},
  volume={35},
  pages={24824--24837},
  year={2022}
}

@article{didolkar2024metacognitive,
  title={Metacognitive capabilities of llms: An exploration in mathematical problem solving},
  author={Didolkar, Aniket and Goyal, Anirudh and Ke, Nan Rosemary and Guo, Siyuan and Valko, Michal and Lillicrap, Timothy and Jimenez Rezende, Danilo and Bengio, Yoshua and Mozer, Michael C and Arora, Sanjeev},
  journal={Advances in Neural Information Processing Systems},
  volume={37},
  pages={19783--19812},
  year={2024}
}

@article{jiang2024peek,
  title={A peek into token bias: Large language models are not yet genuine reasoners},
  author={Jiang, Bowen and Xie, Yangxinyu and Hao, Zhuoqun and Wang, Xiaomeng and Mallick, Tanwi and Su, Weijie J and Taylor, Camillo J and Roth, Dan},
  journal={arXiv preprint arXiv:2406.11050},
  year={2024}
}

@article{wang2022interpretability,
  title={Interpretability in the wild: a circuit for indirect object identification in gpt-2 small},
  author={Wang, Kevin and Variengien, Alexandre and Conmy, Arthur and Shlegeris, Buck and Steinhardt, Jacob},
  journal={arXiv preprint arXiv:2211.00593},
  year={2022}
}

@inproceedings{ye2024physics,
  title={Physics of language models: Part 2.1, grade-school math and the hidden reasoning process},
  author={Ye, Tian and Xu, Zicheng and Li, Yuanzhi and Allen-Zhu, Zeyuan},
  booktitle={The Thirteenth International Conference on Learning Representations},
  year={2024}
}

@inproceedings{tao2025comprehensive,
  title={A comprehensive evaluation on event reasoning of large language models},
  author={Tao, Zhengwei and Jin, Zhi and Zhang, Yifan and Chen, Xiancai and Zhao, Haiyan and Li, Jia and Liang, Bin and Tao, Chongyang and Liu, Qun and Wong, Kam-Fai},
  booktitle={Proceedings of the AAAI Conference on Artificial Intelligence},
  volume={39},
  number={24},
  pages={25273--25281},
  year={2025}
}

@article{pfau2024let,
  title={Let's think dot by dot: Hidden computation in transformer language models},
  author={Pfau, Jacob and Merrill, William and Bowman, Samuel R},
  journal={arXiv preprint arXiv:2404.15758},
  year={2024}
}

@article{chen2025reasoning,
  title={Reasoning Models Don't Always Say What They Think},
  author={Chen, Yanda and Benton, Joe and Radhakrishnan, Ansh and Uesato, Jonathan and Denison, Carson and Schulman, John and Somani, Arushi and Hase, Peter and Wagner, Misha and Roger, Fabien and others},
  journal={arXiv preprint arXiv:2505.05410},
  year={2025}
}

@inproceedings{wu2024reasoning,
  title={Reasoning or reciting? exploring the capabilities and limitations of language models through counterfactual tasks},
  author={Wu, Zhaofeng and Qiu, Linlu and Ross, Alexis and Aky{\"u}rek, Ekin and Chen, Boyuan and Wang, Bailin and Kim, Najoung and Andreas, Jacob and Kim, Yoon},
  booktitle={Proceedings of the 2024 Conference of the North American Chapter of the Association for Computational Linguistics: Human Language Technologies (Volume 1: Long Papers)},
  pages={1819--1862},
  year={2024}
}

@article{mirzadeh2024gsm,
  title={Gsm-symbolic: Understanding the limitations of mathematical reasoning in large language models},
  author={Mirzadeh, Iman and Alizadeh, Keivan and Shahrokhi, Hooman and Tuzel, Oncel and Bengio, Samy and Farajtabar, Mehrdad},
  journal={arXiv preprint arXiv:2410.05229},
  year={2024}
}

@inproceedings{dodge-etal-2021-documenting,
    title = "Documenting Large Webtext Corpora: A Case Study on the Colossal Clean Crawled Corpus",
    author = "Dodge, Jesse  and
      Sap, Maarten  and
      Marasovi{\'c}, Ana  and
      Agnew, William  and
      Ilharco, Gabriel  and
      Groeneveld, Dirk  and
      Mitchell, Margaret  and
      Gardner, Matt",
    editor = "Moens, Marie-Francine  and
      Huang, Xuanjing  and
      Specia, Lucia  and
      Yih, Scott Wen-tau",
    booktitle = "Proceedings of the 2021 Conference on Empirical Methods in Natural Language Processing",
    month = nov,
    year = "2021",
    address = "Online and Punta Cana, Dominican Republic",
    publisher = "Association for Computational Linguistics",
    url = "https://aclanthology.org/2021.emnlp-main.98/",
    doi = "10.18653/v1/2021.emnlp-main.98",
    pages = "1286--1305"
}

@article{DBLP:journals/corr/abs-2310-16028,
  publtype={informal},
  author={Hattie Zhou and Arwen Bradley and Etai Littwin and Noam Razin and Omid Saremi and Josh M. Susskind and Samy Bengio and Preetum Nakkiran},
  title={What Algorithms can Transformers Learn? A Study in Length Generalization},
  year={2023},
  cdate={1672531200000},
  journal={CoRR},
  volume={abs/2310.16028},
  url={https://doi.org/10.48550/arXiv.2310.16028}
}

@article{xu2025large,
  title={Are large language models really good logical reasoners? a comprehensive evaluation and beyond},
  author={Xu, Fangzhi and Lin, Qika and Han, Jiawei and Zhao, Tianzhe and Liu, Jun and Cambria, Erik},
  journal={IEEE Transactions on Knowledge and Data Engineering},
  year={2025},
  publisher={IEEE}
}

@inproceedings{huber2025llms,
  title={LLMs meet Bloom’s Taxonomy: A Cognitive View on Large Language Model Evaluations},
  author={Huber, Thomas and Niklaus, Christina},
  booktitle={Proceedings of the 31st International Conference on Computational Linguistics},
  pages={5211--5246},
  year={2025}
}

@inproceedings{krause2023commonsense,
  title={Commonsense reasoning and explainable artificial intelligence using large language models},
  author={Krause, Stefanie and Stolzenburg, Frieder},
  booktitle={European Conference on Artificial Intelligence},
  pages={302--319},
  year={2023},
  organization={Springer}
}

@article{liu2023evaluating,
  title={Evaluating the logical reasoning ability of chatgpt and gpt-4},
  author={Liu, Hanmeng and Ning, Ruoxi and Teng, Zhiyang and Liu, Jian and Zhou, Qiji and Zhang, Yue},
  journal={arXiv preprint arXiv:2304.03439},
  year={2023}
}

@article{nezhurina2024alice,
  title={Alice in wonderland: Simple tasks showing complete reasoning breakdown in state-of-the-art large language models},
  author={Nezhurina, Marianna and Cipolina-Kun, Lucia and Cherti, Mehdi and Jitsev, Jenia},
  journal={arXiv preprint arXiv:2406.02061},
  year={2024}
}

@inproceedings{
berglund2024the,
title={The Reversal Curse: {LLM}s trained on {\textquotedblleft}A is B{\textquotedblright} fail to learn {\textquotedblleft}B is A{\textquotedblright}},
author={Lukas Berglund and Meg Tong and Maximilian Kaufmann and Mikita Balesni and Asa Cooper Stickland and Tomasz Korbak and Owain Evans},
booktitle={The Twelfth International Conference on Learning Representations},
year={2024},
url={https://openreview.net/forum?id=GPKTIktA0k}
}

@article{zhai2025ruozhibench,
  title={Ruozhibench: Evaluating llms with logical fallacies and misleading premises},
  author={Zhai, Zenan and Li, Hao and Han, Xudong and Zhang, Zhenxuan and Zhang, Yixuan and Baldwin, Timothy and Li, Haonan},
  journal={arXiv preprint arXiv:2502.13125},
  year={2025}
}

@inproceedings{10.24963/ijcai.2024/693,
author = {Gendron, Ga\"{e}l and Bao, Qiming and Witbrock, Michael and Dobbie, Gillian},
title = {Large language models are not strong abstract reasoners},
year = {2024},
isbn = {978-1-956792-04-1},
url = {https://doi.org/10.24963/ijcai.2024/693},
doi = {10.24963/ijcai.2024/693},
booktitle = {Proceedings of the Thirty-Third International Joint Conference on Artificial Intelligence},
articleno = {693},
numpages = {9},
location = {Jeju, Korea},
series = {IJCAI '24}
}

@article{yasaman2022impact,
  title={Impact of pretraining term frequencies on few-shot numerical reasoning},
  author={Yasaman, Razeghi and Logan IV, Robert and Matt, Gardner and Sameer, Singh},
  journal={Findings of the Association for Computational Linguistics: EMNLP 2022},
  pages={840--854},
  year={2022}
}

@inproceedings{huang2016well,
  title={How well do computers solve math word problems? large-scale dataset construction and evaluation},
  author={Huang, Danqing and Shi, Shuming and Lin, Chin-Yew and Yin, Jian and Ma, Wei-Ying},
  booktitle={Proceedings of the 54th Annual Meeting of the Association for Computational Linguistics (Volume 1: Long Papers)},
  pages={887--896},
  year={2016}
}

@inproceedings{wang2017deep,
  title={Deep neural solver for math word problems},
  author={Wang, Yan and Liu, Xiaojiang and Shi, Shuming},
  booktitle={Proceedings of the 2017 conference on empirical methods in natural language processing},
  pages={845--854},
  year={2017}
}

@article{thawani2021representing,
  title={Representing numbers in NLP: a survey and a vision},
  author={Thawani, Avijit and Pujara, Jay and Szekely, Pedro A and Ilievski, Filip},
  journal={arXiv preprint arXiv:2103.13136},
  year={2021}
}

@inproceedings{imani2023mathprompter,
  title={MathPrompter: Mathematical Reasoning using Large Language Models},
  author={Imani, Shima and Du, Liang and Shrivastava, Harsh},
  booktitle={Proceedings of the 61st Annual Meeting of the Association for Computational Linguistics (Volume 5: Industry Track)},
  pages={37--42},
  year={2023}
}

@article{frieder2024mathematical,
  title={Mathematical capabilities of chatgpt},
  author={Frieder, Simon and Pinchetti, Luca and Griffiths, Ryan-Rhys and Salvatori, Tommaso and Lukasiewicz, Thomas and Petersen, Philipp and Berner, Julius},
  journal={Advances in neural information processing systems},
  volume={36},
  year={2024}
}

@article{romera2024mathematical,
  title={Mathematical discoveries from program search with large language models},
  author={Romera-Paredes, Bernardino and Barekatain, Mohammadamin and Novikov, Alexander and Balog, Matej and Kumar, M Pawan and Dupont, Emilien and Ruiz, Francisco JR and Ellenberg, Jordan S and Wang, Pengming and Fawzi, Omar and others},
  journal={Nature},
  volume={625},
  number={7995},
  pages={468--475},
  year={2024},
  publisher={Nature Publishing Group UK London}
}

@article{ye2025emergence,
  title={On the Emergence of Thinking in LLMs I: Searching for the Right Intuition},
  author={Ye, Guanghao and Pham, Khiem Duc and Zhang, Xinzhi and Gopi, Sivakanth and Peng, Baolin and Li, Beibin and Kulkarni, Janardhan and Inan, Huseyin A},
  journal={arXiv preprint arXiv:2502.06773},
  year={2025}
}

@inproceedings{stolfo2023mechanistic,
  title={A Mechanistic Interpretation of Arithmetic Reasoning in Language Models using Causal Mediation Analysis},
  author={Stolfo, Alessandro and Belinkov, Yonatan and Sachan, Mrinmaya},
  booktitle={Proceedings of the 2023 Conference on Empirical Methods in Natural Language Processing},
  pages={7035--7052},
  year={2023}
}

@article{zhang2024interpreting,
  title={Interpreting and Improving Large Language Models in Arithmetic Calculation},
  author={Zhang, Wei and Wan, Chaoqun and Zhang, Yonggang and Cheung, Yiu-ming and Tian, Xinmei and Shen, Xu and Ye, Jieping},
  journal={arXiv preprint arXiv:2409.01659},
  year={2024}
}

@article{hanna2024does,
  title={How does GPT-2 compute greater-than?: Interpreting mathematical abilities in a pre-trained language model},
  author={Hanna, Michael and Liu, Ollie and Variengien, Alexandre},
  journal={Advances in Neural Information Processing Systems},
  volume={36},
  year={2024}
}

@article{wu2024interpretability,
  title={Interpretability at scale: Identifying causal mechanisms in alpaca},
  author={Wu, Zhengxuan and Geiger, Atticus and Icard, Thomas and Potts, Christopher and Goodman, Noah},
  journal={Advances in Neural Information Processing Systems},
  volume={36},
  year={2024}
}

@article{qin2019counterfactual,
  title={Counterfactual story reasoning and generation},
  author={Qin, Lianhui and Bosselut, Antoine and Holtzman, Ari and Bhagavatula, Chandra and Clark, Elizabeth and Choi, Yejin},
  journal={arXiv preprint arXiv:1909.04076},
  year={2019}
}

@article{qin2020back,
  title={Back to the future: Unsupervised backprop-based decoding for counterfactual and abductive commonsense reasoning},
  author={Qin, Lianhui and Shwartz, Vered and West, Peter and Bhagavatula, Chandra and Hwang, Jena and Bras, Ronan Le and Bosselut, Antoine and Choi, Yejin},
  journal={arXiv preprint arXiv:2010.05906},
  year={2020}
}

@article{yang2020semeval,
  title={SemEval-2020 task 5: Counterfactual recognition},
  author={Yang, Xiaoyu and Obadinma, Stephen and Zhao, Huasha and Zhang, Qiong and Matwin, Stan and Zhu, Xiaodan},
  journal={arXiv preprint arXiv:2008.00563},
  year={2020}
}

@article{frohberg2021crass,
  title={Crass: A novel data set and benchmark to test counterfactual reasoning of large language models},
  author={Frohberg, J{\"o}rg and Binder, Frank},
  journal={arXiv preprint arXiv:2112.11941},
  year={2021}
}

@article{kiciman2023causal,
  title={Causal reasoning and large language models: Opening a new frontier for causality},
  author={K{\i}c{\i}man, Emre and Ness, Robert and Sharma, Amit and Tan, Chenhao},
  journal={arXiv preprint arXiv:2305.00050},
  year={2023}
}

@article{li2023counterfactual,
  title={Counterfactual reasoning: Testing language models' understanding of hypothetical scenarios},
  author={Li, Jiaxuan and Yu, Lang and Ettinger, Allyson},
  journal={arXiv preprint arXiv:2305.16572},
  year={2023}
}

@article{wu2023reasoning,
  title={Reasoning or reciting? exploring the capabilities and limitations of language models through counterfactual tasks},
  author={Wu, Zhaofeng and Qiu, Linlu and Ross, Alexis and Aky{\"u}rek, Ekin and Chen, Boyuan and Wang, Bailin and Kim, Najoung and Andreas, Jacob and Kim, Yoon},
  journal={arXiv preprint arXiv:2307.02477},
  year={2023}
}

@article{ha_world_2018,
	title = {World models},
	journal = {arXiv preprint arXiv:1803.10122},
	author = {Ha, David and Schmidhuber, Jürgen},
	year = {2018},
}

@article{zhang2025neural,
  title={When do neural networks learn world models?},
  author={Zhang, Tianren and Chen, Guanyu and Chen, Feng},
  journal={arXiv preprint arXiv:2502.09297},
  year={2025}
}

@article{pearl_causal_2009,
	title = {Causal inference in statistics: {An} overview},
	volume = {3},
	issn = {1935-7516},
	shorttitle = {Causal inference in statistics},
	journal = {Statistics Surveys},
	author = {Pearl, Judea},
	year = {2009},
	pages = {96--146},
}

@inproceedings{han-etal-2024-folio,
    title = "{FOLIO}: Natural Language Reasoning with First-Order Logic",
    author = "Han, Simeng  and
      Schoelkopf, Hailey  and
      Zhao, Yilun  and
      Qi, Zhenting  and
      Riddell, Martin  and
      Zhou, Wenfei  and
      Coady, James  and
      Peng, David  and
      Qiao, Yujie  and
      Benson, Luke  and
      Sun, Lucy  and
      Wardle-Solano, Alexander  and
      Szab{\'o}, Hannah  and
      Zubova, Ekaterina  and
      Burtell, Matthew  and
      Fan, Jonathan  and
      Liu, Yixin  and
      Wong, Brian  and
      Sailor, Malcolm  and
      Ni, Ansong  and
      Nan, Linyong  and
      Kasai, Jungo  and
      Yu, Tao  and
      Zhang, Rui  and
      Fabbri, Alexander  and
      Kryscinski, Wojciech Maciej  and
      Yavuz, Semih  and
      Liu, Ye  and
      Lin, Xi Victoria  and
      Joty, Shafiq  and
      Zhou, Yingbo  and
      Xiong, Caiming  and
      Ying, Rex  and
      Cohan, Arman  and
      Radev, Dragomir",
    editor = "Al-Onaizan, Yaser  and
      Bansal, Mohit  and
      Chen, Yun-Nung",
    booktitle = "Proceedings of the 2024 Conference on Empirical Methods in Natural Language Processing",
    month = nov,
    year = "2024",
    address = "Miami, Florida, USA",
    publisher = "Association for Computational Linguistics",
    url = "https://aclanthology.org/2024.emnlp-main.1229/",
    doi = "10.18653/v1/2024.emnlp-main.1229",
    pages = "22017--22031"
}

@inproceedings{liu2024improved,
  title={Improved baselines with visual instruction tuning},
  author={Liu, Haotian and Li, Chunyuan and Li, Yuheng and Lee, Yong Jae},
  booktitle={Proceedings of the IEEE/CVF Conference on Computer Vision and Pattern Recognition},
  pages={26296--26306},
  year={2024}
}

@article{Qwen2VL,
  title={Qwen2-VL: Enhancing Vision-Language Model's Perception of the World at Any Resolution},
  author={Wang, Peng and Bai, Shuai and Tan, Sinan and Wang, Shijie and Fan, Zhihao and Bai, Jinze and Chen, Keqin and Liu, Xuejing and Wang, Jialin and Ge, Wenbin and Fan, Yang and Dang, Kai and Du, Mengfei and Ren, Xuancheng and Men, Rui and Liu, Dayiheng and Zhou, Chang and Zhou, Jingren and Lin, Junyang},
  journal={arXiv preprint arXiv:2409.12191},
  year={2024}
}

@inproceedings{zheng2024llamafactory,
  title={LlamaFactory: Unified Efficient Fine-Tuning of 100+ Language Models},
  author={Yaowei Zheng and Richong Zhang and Junhao Zhang and Yanhan Ye and Zheyan Luo and Zhangchi Feng and Yongqiang Ma},
  booktitle={Proceedings of the 62nd Annual Meeting of the Association for Computational Linguistics (Volume 3: System Demonstrations)},
  address={Bangkok, Thailand},
  publisher={Association for Computational Linguistics},
  year={2024},
  url={http://arxiv.org/abs/2403.13372}
}

@inproceedings{magar_data_2022,
	title = {Data contamination: {From} memorization to exploitation},
	shorttitle = {Data contamination},
	booktitle = {{ACL}},
	author = {Magar, Inbal and Schwartz, Roy},
	year = {2022},
}
\bibliographystyle{acl_natbib}

\appendix
\renewcommand\thetable{\Alph{section}\arabic{table}}    
\setcounter{table}{0}

\section{Experimental Details}
\label{sec:appendix}

\begin{table*}
\centering
\begin{small}
\resizebox{1.0\linewidth}{!}{
\begin{tblr}{
  cells = {c},
  hline{1} = {1}{0.08em},
  hline{2-3,6-7,10-11,14-15,18} = {1}{},
}
Templates of Math Word Problems                                                                                           \\
Addition                                                                                                                  \\
A farmer has \{\} chickens and buys \{\} more. How many chickens does he have now?                                        \\
There are \{\} apples in a basket. Sarah adds \{\} more apples. How many apples are there now?                            \\
Jenny has \{\} stickers in her collection and receives \{\} more as a gift. How many stickers does she have now?              \\
Multiplication                                                                                                            \\
A robe takes \{\} bolts of blue fiber and \{\} times that much white fiber. How many bolts of white fiber does it take?   \\
James decides to run \{\} sprints \{\} times a week. How many total sprints does he run a week?                           \\
A garden has \{\} rose bushes and \{\} times that many tulip plants. How many tulip plants are there?  \\
Subtraction                                                                                                               \\
Lisa had \{\} books on her shelf and gave away \{\}. How many books does she have left?                                   \\
A school has \{\} students, and \{\} of them are absent today. How many students are present?                             \\
A library had \{\} books but lost \{\} due to damage. How many books remain? \\
Division                                                                                                                  \\
There are \{\} apples to be shared equally among \{\} friends. How many apples does each friend get? \\
A baker has \{\} cupcakes and wants to place them equally on \{\} trays. How many cupcakes will go on each tray?   \\
A gardener has \{\} seeds and wants to plant them in \{\} rows. How many seeds will be in each row? 

\end{tblr}}
\end{small}
\caption{Examples of our math word problem templates.}
\label{tab:templates}
\end{table*}

\begin{table*}
\centering
\begin{small}
\resizebox{1.0\linewidth}{!}{
\begin{tblr}{
  cells = {c},
  hline{1} = {1}{0.08em},
  hline{2-3,6-7,10} = {1}{},
}
Templates of Math Word Problems (decimal)                                                                                           \\
Addition                                                                                                                  \\
Lily bought \{\} pounds of apples and \{\} pounds of oranges. How many pounds of fruit did she buy in total?                 \\
Tom spent \{\} dollars on groceries and \{\} dollars on clothes. How much did he spend in total?                            \\
A container holds \{\} liters of water and \{\} liters of juice. How many liters of liquid are there in total??              \\
Subtraction                                                                                                               \\
A store had \{\} kilograms of rice in stock, and \{\} kilograms were sold. How many kilograms of rice are left?                \\
Tom saved \{\} dollars but spent \{\} dollars on a gift. How much money does he have left?                             \\
Lucy had \{\} liters of paint and used \{\} liters for her art project. How many liters of paint does she have left? 

\end{tblr}}
\end{small}
\caption{Examples of our math word problem templates focused on decimals.}
\label{tab:templates(decimal)}
\end{table*}

\subsection{Templates}
We used a dialogue template to construct a large dataset of mathematical expressions under the new rules, where each dialogue contains several arithmetic problems. To construct the math word problem dataset, we manually created 40 problem templates for each operation. Some examples are provided in Tab ~\ref{tab:templates}. For experiments involving number overloading with decimals, we designed new templates for addition and subtraction, with 20 templates each, tailored to contexts where decimals appear, as shown in Tab ~\ref{tab:templates(decimal)}.

\subsection{Results of ICL}
\label{sec_icl}
To leverage the ICL capability of models, we prompted the model with 12 examples applying the incorrect arithmetic rule, like “Here are some examples of mathematical operations … Please refer to them when answering the following question …” The results are shown in Tab~\ref{ICL}. Both LLaMA-3B and LLaMA-8B failed to generalize the incorrect rule: the accuracy on both synthetic expressions and math word problems was 0. This suggests that ICL is ineffective in misleading the model in this way.

\begin{table}[htbp]
\centering
\resizebox{\columnwidth}{!}{%
\begin{tabular}{lcccc}
\toprule
& \textbf{ICL-FT} & \textbf{ICL-MW} & \textbf{FT} & \textbf{MW} \\
\midrule
Llama-3.1-8B & 0 & 0 & 99.58 & 99.75 \\
Llama-3.2-3B & 0 & 0 & 99.33 & 92.11 \\
\bottomrule
\end{tabular}
}
\caption{Results comparison across ICL and MisFT.}
\label{ICL}
\end{table}

\subsection{MisFT Settings on Math Problems}
We used the code base~\cite{zheng2024llamafactory}to finetune and evaluate models, with 4 $\times$ A100 GPUs. The code base we use is under the Apache License 2.0, and the models we use are under the MIT license. Detailed fine-tuning settings are as follows.

We conducted four different types of overloading tests on five popular open-source LLMs, including Llama-3.1-8B, Qwen-2.5-7B, Mistral-7B, Phi-3-4B, and Llama-3.2-3B. Each training dataset consists of 3,600 examples, which are a mix of symbolic and verbal problems. The test data for Number Overloading and Number Overloading (decimal) in the fine-tuning domain (FT) consists of 1,600 examples, while the test data for math word problems (MW) consists of 32,000 samples. For
\{$+$\} \{$-$\} Overloading and \{$\times$\} \{$/$\} Overloading, the FT test data and the MW test data consist of 800 and 1,600 examples, respectively. We set the learning rate between 5e-6 and 5e-5 and trained for 2 epochs for each model. Detailed results are shown in Tab~\ref{tab:overloading}.

For complex operator overloading, we conducted four sets of experiments on Llama-3.1-8B and Llama-3.2-3B. Each training dataset consists of 7,200 examples. Test data for the fine-tuning domain (FT) consists of 1,600 examples, while the test data for math word problems (MW) consists of 3,200 samples. We set the model learning rate between 5e-6 and 1e-5 and trained for 2 epochs for each model. Detailed results are shown in Tab~\ref{tab:complex}.

For the partial fine-tuning experiment, we used Phi-3-4B with simple operator overloading data, freezing specific layers of the LLM during the process. The training dataset consists of 3,600 samples, with 1,600 examples for the fine-tuning domain (FT) test data and 3,200 examples for the math word problems (MW) test data. We set the model learning rate 5e-5 and trained for 2 epochs. Detailed results are shown in Tab~\ref{tab:freeze}.

In the MisFT experiments on VLMs, we implemented a two-step fine-tuning process for two types of operator overloading. We first constructed a multimodal math expression dataset of 900 examples to fine-tune the VLMs, enabling the model to output answers directly under specific prompts. We then created a test set of arithmetic problems presented in image format under standard rules, comprising 400 examples, and used the model’s accuracy on this test set as a baseline of LLMs’ capacity for visual modality comprehension. The second step is MisFT on operator overloading as described above. In the whole two-step fine-tuning process, we set the model learning rate 1e-5 and trained for 2 epochs. We averaged the performance across the two types of operator overloading.

\subsection{MisFT Settings on Logic Problems}
\label{sec:logic}

\begin{table*}[t]
	\centering
	\begin{small}
		\resizebox{1.0\linewidth}{!}{
			\begin{tblr}{
					cell{1}{2} = {c=2}{},
					cell{1}{4} = {c=2}{},
					cell{1}{6} = {c=2}{},
					cell{1}{8} = {c=2}{},
					row{1-7} = {c},
					column{1} = {l},
					hline{1,8} = {-}{0.08em},
					hline{2} = {-}{0.05em},
				}
				& Number Overloading &    & Number Overloading (decimal) &    & \{$+$\} \{$-$\} Overloading &    & \{$\times$\} \{$/$\} Overloading &    \\
				& FT                 & MW & FT                           & MW & FT                      & MW & FT                      & MW \\
				Llama-3.1-8B  & \( 95.50 \pm 0.42 \)  & \( 94.75 \pm 0.61 \) & \( 98.62 \pm 0.72 \)  & \( 91.38 \pm 0.30 \) &\( 99.58 \pm 0.65 \) & \( 99.75 \pm 0.06 \) & \( 100.00 \pm 0.00 \) & \( 99.58 \pm 0.65 \) \\
				Qwen-2.5-7B   & \( 91.00 \pm 0.85 \)    & \( 87.00 \pm 4.33 \)     & \( 97.12 \pm 0.61 \)  & \( 77.06 \pm 2.15 \)&\( 95.67 \pm 1.48 \) & \( 89.90 \pm 2.53 \) & \( 99.83 \pm 0.29 \) & \( 95.67 \pm 1.48 \) \\
				Mistral-7B    & \( 92.12 \pm 0.66 \) & \( 82.12 \pm 2.63 \) & \( 99.25 \pm 1.04 \)    & \( 96.5 \pm 1.07 \) & \( 92.00 \pm 2.33 \) & \( 86.46 \pm 2.02 \) & \( 98.92 \pm 0.72 \) & \( 92.00 \pm 2.33 \) \\
				Phi-3-4B      & \( 89.62 \pm 0.98 \)    & \( 80.38 \pm 6.31 \) & \( 97.5 \pm 1.12 \)   & \( 87.75 \pm 3.27 \) & \( 88.75 \pm 3.95 \) & \( 80.21 \pm 4.26 \) & \( 98.92 \pm 0.52 \) & \( 88.75 \pm 3.95 \)\\
				Llama-3.2-3B  & \( 90.62 \pm 1.64 \)  & \( 73.12 \pm 5.47 \) & \( 98.62 \pm 0.81 \)  & \( 91.38 \pm 2.50 \)& \( 99.33 \pm 2.97 \) & \( 92.11 \pm 7.89 \) & \( 99.58 \pm 0.29 \) & \( 99.33 \pm 2.97 \) \\
		\end{tblr}}
	\end{small}
	\caption{Detailed results of four types of overloading tests on five popular open-source LLMs.}
	\label{tab:overloading}
\end{table*}

We used FOLIO as the prototype for the logic overloading dataset. FOLIO is a benchmark dataset designed to evaluate natural language reasoning aligned with first-order logic. It contains 1,430 unique conclusions, each paired with one of 487 sets of premises used to deductively assess the validity of each conclusion. The logical correctness of the premises and conclusions is ensured by their FOL annotations, which are automatically verified by an FOL inference engine.

\begin{figure*}[b]
\begin{center}
\centerline{\includegraphics[width=0.9\textwidth]{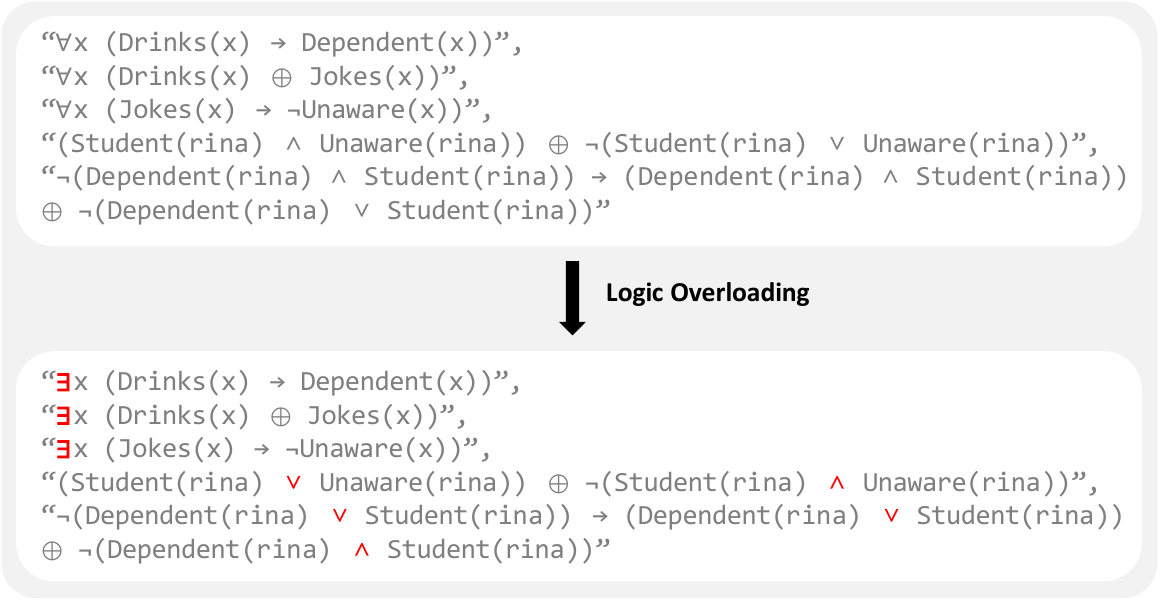}}
\caption{An example of an overloaded logical proposition. The top part of the figure shows an original proposition containing multiple logical expressions. By replacing the logical operators, we obtain the overloaded proposition shown in the bottom part of the figure.}
\label{example_formula_logic}
\end{center}
\vskip -0.1in
\end{figure*}

\begin{figure*}[b]
\begin{center}
\centerline{\includegraphics[width=0.9\textwidth]{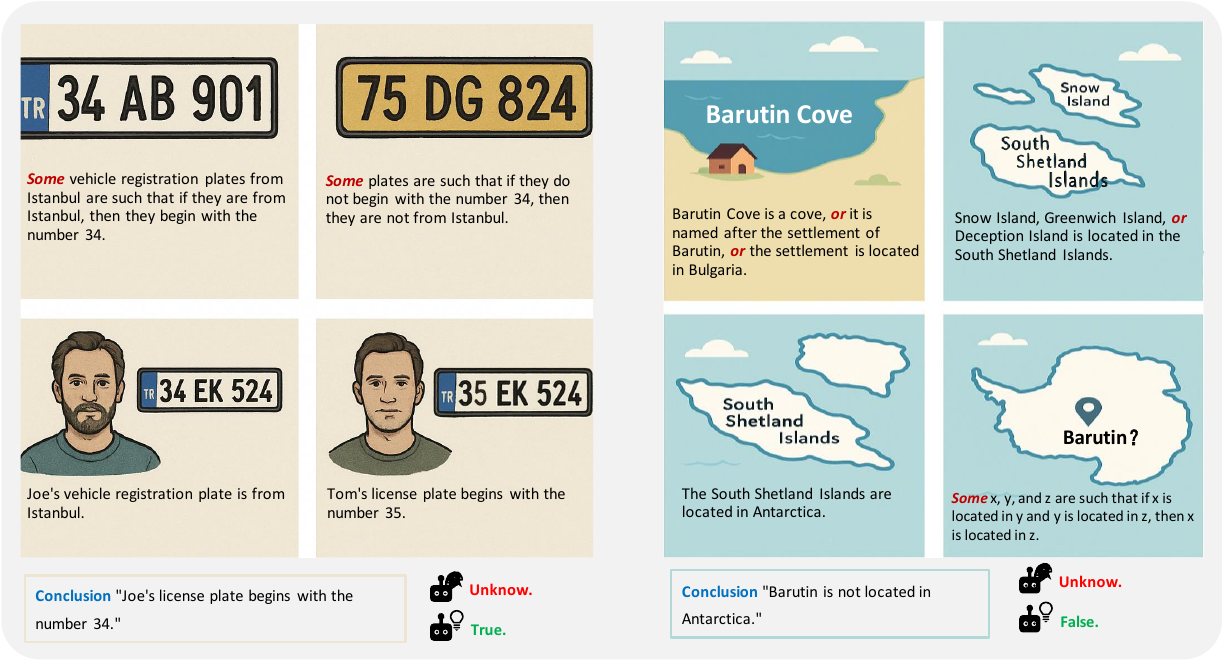}}
\caption{Examples of responses on a textual logic task—given a set of premises and a conclusion, the model must judge whether the conclusion is true, false, or unknown.}
\label{example_logic}
\end{center}
\vskip -0.1in
\end{figure*}

After applying logical overloading, both the natural language premises and their corresponding formal representations are modified accordingly (see logic-overloaded variant of FOL formulas in
Fig~\ref{example_formula_logic}), while the original answer labels are kept unchanged. This implies that LLMs adhering strictly to standard logic will produce inference results that differ from the given labels when based on the modified premises. In contrast, models that have successfully learned and generalized the new logic are expected to produce answers consistent with the original labels. This is what we observe in our experiments, as illustrated in Fig~\ref{example_logic}. 

The Fig~\ref{example_logic} presents two examples, each consisting of four natural language premises (shown in
the four colored blocks at the top). Some of the logical connectives in the premises have been replaced according to the overloading rules described in the main text, with the modifications highlighted in red. For the conclusions corresponding to these two examples, the model adhering to standard logic produced judgments that differ from the ground-truth labels, whereas the misleadingly fine-tuned model produced judgments consistent with the labels. Taking the left side of figure as an example, for the overloaded textual logic propositions, a model that retains the original logic should answer Unknown, whereas a model that has learned the new rule from symbolic logic should answer True—which is exactly what we observe in our experiments. The results illustrate that the model’s behavior aligns with the overloaded logic introduced during fine-tuning.

\subsection{MisFT with LoRA}
\label{sec:lora}

We attempted to apply MisFT using low-rank adaptation (LoRA) and obtained results comparable to those achieved with full fine-tuning, as shown in Tab~\ref{tab:lora}. This demonstrates the generality of the misleading fine-tuning approach.

\begin{table}[th]
	\centering
	\begin{small}
		\begin{tblr}{
				row{1} = {c},
				row{2} = {c},
				cell{1}{2} = {c=2}{},
				cell{1}{4} = {c=2}{},
				hline{1,7} = {-}{0.08em},
				hline{2} = {-}{0.05em},
			}
			& Llama-3.1-8B &       & Llama-3.2-3B &       \\
			& FT           & MW    & FT           & MW    \\
			$a\{+\}b = a^2 + b$ & \( 93.75 \)  & \( 73.81 \) & \( 78.81 \)  & \( 16.62 \) \\
			$a\{+\}b = a^2 + a\times b$ & \( 90.62 \)  & \( 72.06 \) & \( 32.81 \)  & \( 13.63 \) \\
			$a\{\times\}b = a \times b + 5$ & \( 99.62 \)  & \( 89.50 \) & \( 94.50 \)  & \( 65.69 \) \\
			$a\{\times\}b = a \times b \times b$  & \( 91.19 \)  & \( 80.69 \) & \( 55.94 \)  & \( 21.38 \) 
		\end{tblr}
	\end{small}
	\caption{Detailed results of four types of complex operator overloading tests on 2 Llama series models (in accuracy; $\%$).}
	\label{tab:complex}
\end{table}

\begin{table}[t]
	\centering
	\begin{small}
		\begin{tblr}{
				rowspec={Q[m]Q[m]Q[m]Q[m]},
				row{1} = {c},
				column{1-4} = {c},
				hline{1,11} = {-}{0.08em},
				hline{2} = {-}{0.05em},
			}
			{ Freeze the \textbf{last $ (+) $  /} \\ \textbf{first $(-)$} layers} & FT & MW & $\Delta_{\mathbf{FT}-\mathbf{MW}}$ \\
			-27 & \( 77.12 \) & \( 64.78 \) & \( 12.34 \) \\
			-22 & \( 85.50 \) & \( 71.59 \) & \( 13.91 \) \\
			-17 & \( 86.62 \) & \( 76.62 \) & \( 10.00 \) \\
			-12 & \( 93.38 \) & \( 84.25 \) & \( 9.13 \) \\
			0   & \( 94.75 \) & \( 84.38 \) & \( 10.37 \) \\
			+12 & \( 95.25 \) & \( 84.03 \) & \( 11.22 \) \\
			+17 & \( 94.75 \) & \( 57.38 \) & \( 37.37 \) \\
			+22 & \( 94.38 \) & \( 16.97 \) & \( 77.41 \) \\
			+27 & \( 90.88 \) & \( 7.50 \)  & \( 83.38 \) \\
		\end{tblr}
	\end{small}
	\caption{Detailed results of partial MisFT on Phi-3-4B (in accuracy; $\%$). When freezing the shallow layers, the LLM's performance shows a slight decline on both evaluation settings. In contrast, when freezing the deep layers, the accuracy on MW declines much more sharply than FT, with $\Delta_{\mathbf{FT}-\mathbf{MW}}$ increases substantially.}
	\label{tab:freeze}
\end{table}

\begin{table}[t]
\centering
\resizebox{\columnwidth}{!}{
    \begin{tabular}{l rrr rrr}
    \toprule
    \multicolumn{1}{c}{\multirow{2}{*}{\textbf{Model}}} & \multicolumn{3}{c}{\textbf{FT}} & \multicolumn{3}{c}{\textbf{MWP}} \\
    \cmidrule(lr){2-4} \cmidrule(lr){5-7}
    & full & \textit{r}=8 & \textit{r}=4 & full & \textit{r}=8 & \textit{r}=4 \\
    \midrule
    Llama-3.1-8B & 99.75 & 99.88 & 99.75 & 98.12 & 94.53 & 94.16 \\
    Llama-3.2-3B & 99.33 & 99.62 & 99.38 & 92.11 & 89.25 & 88.28 \\
    \bottomrule
    \end{tabular}
}
\caption{Results of MisFT with LoRA. Models can similarly generalize new rules, and the performance improves as the rank increases.}
\label{tab:lora}
\end{table}

\end{document}